\documentclass[10pt,twocolumn,letterpaper]{article}

 \usepackage[pagenumbers]{cvis} % To force page numbers, e.g. for an arXiv version

\usepackage{xspace}
\DeclareMathOperator*{\argmax}{arg\,max}

\usepackage[all]{nowidow}

\usepackage{dsfont}

\definecolor{cvisblue}{rgb}{0.21,0.49,0.74}
\usepackage[pagebackref,breaklinks,colorlinks,allcolors=cvisblue]{hyperref}

\def\confName{CVIS}
\def\confYear{2025}

\title{Effects of Initialization Biases on Deep Neural Network Training Dynamics}

\author{Nicholas Pellegrino$^{1}$\thanks{Indicates equal contribution, joint first-authorship.}, David Szczecina$^{1,2}$\footnotemark[1], \& Paul Fieguth$^{1}$\\
$^1$Vision and Image Processing Group, Systems Design Engineering, University of Waterloo\\
$^2$Mechanical \& Mechatronics Engineering, University of Waterloo\\
{\tt\small \{npellegr,dszczeci,pfieguth\}@uwaterloo.ca}
}

\begin{document}
\maketitle
\begin{abstract} % Max 150 words
Untrained large neural networks, just after random initialization, tend to favour a small subset of classes, assigning high predicted probabilities to these few classes and approximately zero probability to all others. 
This bias, termed \emph{Initial Guessing Bias}, affects the early training dynamics, when the model is fitting to the coarse structure of the data. 
The choice of loss function against which to train the model has a large impact on how these early dynamics play out. 
Two recent loss functions, Blurry and Piecewise-zero loss, were designed for robustness to label errors but can become unable to steer the direction of training when exposed to this initial bias. 
Results indicate that the choice of loss function has a dramatic effect on the early phase training of networks, and highlights the need for careful consideration of how Initial Guessing Bias may interact with various components of the training scheme.
\end{abstract}    
\section{Introduction}
\label{sec:introduction}

\begin{figure}[t]
    \centering
    \includegraphics[width=\columnwidth]{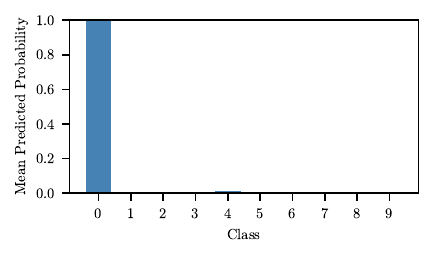}
    \vspace{-2.0em}
    \caption{Predicted probability for each class, averaged over the validation set, directly after model initialization. Observe that one class (class 0, randomly) is \textit{highly} favoured ($\bar{p}_0\approx1$) relative to other classes ($\bar{p}_y\approx0$ for $y\ne0$) as a result of severe Initial Guessing Bias.}
    \label{fig:mean_pred_probs}
    \vspace{-1.0em}
\end{figure}

In the supervised training of deep neural networks, an often-overlooked component of the training is the behaviour of the network before it has been exposed to any labelled data. 
Recent investigations~\cite{francazi2024initial,francazi2025you,bassi2025left} into randomly initialized networks have revealed a surprising and systematic phenomenon: many architectures exhibit a strong preference for a small subset of classes immediately after initialization, assigning high predicted probabilities to these favoured classes and near-zero probabilities to the rest, nearly regardless of the input provided to the model. 
This effect is demonstrated in \Cref{fig:mean_pred_probs}, whereby one class is favoured.
This systematic tendency is termed \textit{Initial Guessing Bias} (IGB)~\cite{francazi2024initial}, which impacts not only the initial distributions but persists in more subtle ways even after training.

IGB most directly impacts models in the earliest stages of training, just after initialization.
During this phase, the loss function plays a particularly critical role: it determines how strongly the model is encouraged to correct its initial biases and how quickly it moves toward a meaningful representation. 
While standard loss functions such as Cross-Entropy (CE)~\cite{good1952rational} provide strong gradients even when predicted probabilities are extremely small, recent work~\cite{zhang2018generalized,ye2023active,ghosh2015making,ghosh2017robust,ma2020normalized,wang2019symmetric,lyucurriculum,zhou2021asymmetric,menon2020can,pang2018towards,pellegrino2024loss} has introduced alternative objectives designed to improve robustness to label errors. 

Two such objectives --- Blurry Loss (BL) and Piecewise-Zero Loss (PZ)~\cite{pellegrino2024loss} --- attenuate or eliminate gradients when the model produces low probabilities for the target class. 
These loss functions were developed to reduce the harmful influence of incorrect labels, examples of which tend to exhibit low predicted probability from models that have generalized, but the behaviour of these losses in the presence of strong initialization-induced biases has not yet been carefully examined.
If a loss function supplies strong gradients for low-probabilities (as CE does), the model can quickly correct these biases, but for losses such as BL and PZ, which produce zero or near-zero gradients on low-probability classes, the network may struggle to overcome its initial bias, leading to slow or stalled early learning.
Thus, the choice of loss function may determine whether a model escapes the basin carved out by IGB or remains trapped during a critical early window of training.

% This interaction between IGB and loss functions is potentially problematic. 
% If a model begins in a state where many classes are assigned extremely small probability due to IGB, then robust losses that provide zero or near-zero gradients in this regime may be unable to meaningfully update the model. 
% In contrast, CE actively counteracts these extreme initial predictions: 
% classes with excessively high probabilities are rapidly pushed down, while 
% suppressed classes receive large corrective gradients. 

This work studies how IGB interacts with CE, BL, and PZ during the earliest phase of training. 
% Firstly, the presence and severity of IGB is characterised, visualizing the initial predicted probability distributions and illustrating how strongly some classes are favoured relative to the others. 
To examine how different loss functions respond to these initial biases, the predicted probabilities, $p(y|x)$, and per-class accuracies are monitored batch-to-batch throughout training. 
CE results in a re-balancing effect whereby the initially dominant class is corrected downward while the suppressed classes rise, until all predicted probabilities tend to move as a group towards higher values. 
BL produces similar outcomes, but at a decreased rate; however, PZ does not allow classes with low predicted probability to contribute meaningfully to training, resulting in the initially favoured class dominating throughout. 

These findings suggest that initialization interacts deeply with the choice of loss function. 
For robust training pipelines, designed to handle label errors, the combination of strong IGB and gradient-suppressing loss functions may lead to extremely slow convergence or complete failure to begin learning. 
These results underscore the importance of understanding how initialization, loss design, and early-stage optimization jointly shape training trajectories, especially in scenarios where robustness is a priority.

\section{Background}
\label{sec:background}

\subsection{Initial Guessing Bias (IGB)}
\label{subsec:bg_igb}
Recent work~\cite{francazi2024initial,francazi2025you,bassi2025left} has shown that untrained neural \mbox{networks} exhibit a systematic and architecture-dependent Initial Guessing Bias (IGB)~\cite{francazi2024initial}. 
Immediately after random initialization, the predicted class distribution is far from uniform: instead, the model tends to assign disproportionately high probability to a small subset of classes while assigning nearly zero probability to the majority. 
This effect is consistent across a range of architectures and arises from subtle asymmetries introduced by weight initialization, activation patterns, and network depth. 
IGB therefore determines the ``starting point'' from which training trajectories begin, and can lead to more subtle biases that persist after training.

\subsection{Loss Functions}
\label{subsec:bg_loss}
In supervised classification, models are typically trained using the Cross-Entropy (CE)~\cite{good1952rational} loss, which encourages predicted probability distributions, $p_y=p(y|x)$, to match one-hot ground truth labels. 
For an input–label pair $(x,y)$, the CE loss is defined as
\begin{equation}
    \text{CE}(p_y) = -\log(p_y).
    \label{eq:ce}
\end{equation}
CE provides large corrective gradients when the predicted probability of the true class is small, making it most sensitive to examples that are not being correctly classified.

Blurry Loss (BL) and Piecewise-Zero Loss (PZ)~\cite{pellegrino2024loss} were introduced to mitigate the influence of mislabelled data by down-weighting or eliminating gradients when the predicted probability of the target class is below a threshold. 

Motivated by Focal Loss~\cite{lin2017focal}, which emphasizes difficult-to-classify examples, \textit{Blurry Loss} was designed to de-emphasize such samples through the inclusion of a multiplicative factor with weighting parameter $\gamma$.
The Blurry Loss is defined as
\begin{equation}
    \mathrm{BL}(p_y) = -\,p_y^\gamma\cdot\log(p_y).
    \label{eq:bl}
\end{equation}
Note that BL has a region of \emph{positive} gradient for $p_{y}<e^{-1/\gamma}$, steering training \emph{against} away from the labelled class if $p_{y}$ is sufficiently low. 
At $\gamma=0$, Blurry Loss is equivalent to Cross Entropy Loss.

\textit{Piecewise-zero Loss} was designed to ignore difficult-to-classify samples (those with low $p_y$) under the assumption that these are likely mislabelled.
If predicted probability is beneath some cutoff, $p_{y}\leq{c\in[0,1]}$, a loss of zero is assigned (with zero gradient, not impacting training).
Piecewise-zero Loss is defined as
\begin{equation}
    \mathrm{PZ}(p_y) = 
    \begin{cases} 
      0 & p_y\leq c, \\
      \mathrm{CE}(p_y) = -\log(p_y) & p_y> c. 
   \end{cases}
   \label{eq:pz}
\end{equation}
Piecewise-zero Loss is equivalent to Cross-Entropy Loss for $c=0$.

\subsection{Addressing Label Errors in Training Data}
\label{subsec:bg_label_errors}
In many real-world datasets, labels are corrupted by human error, automated annotation artifacts, or inherent class ambiguity~\cite{northcutt2021pervasive}. 
Mislabelled data can mislead the model early in training, causing it to over-fit incorrect labels and degrade generalization~\cite{algan2021image,song2022learning,northcutt2021confident}. 
Classical CE loss is known to be sensitive to mislabelled examples because it most heavily penalizes low predicted probabilities on the (possibly incorrect) targets. 
As a result, there has been recent interest~\cite{zhang2018generalized,ye2023active,ghosh2015making,ghosh2017robust,ma2020normalized,wang2019symmetric,lyucurriculum,zhou2021asymmetric,menon2020can,pang2018towards,pellegrino2024loss} in designing robust loss functions, such as BL and PZ~\cite{pellegrino2024loss} described in \Cref{eq:bl,eq:pz} in \Cref{subsec:bg_loss}, that suppress the influence of potentially wrong labels.

\begin{figure*}[tp]
\centering

\begin{tabular}{@{}l c@{}}
% Header row
& \hspace{0.3in} Predicted Probabilities, $\bar{p}_k$ \hspace{1.6in} Per-class Accuracy, $\mathrm{Acc}_k$\\
% ---------- Row 1 ----------
CE &
\begin{minipage}{0.9\textwidth}
    \centering
    \begin{subfigure}[b]{0.49\textwidth}
        \centering
        \includegraphics[width=\linewidth]{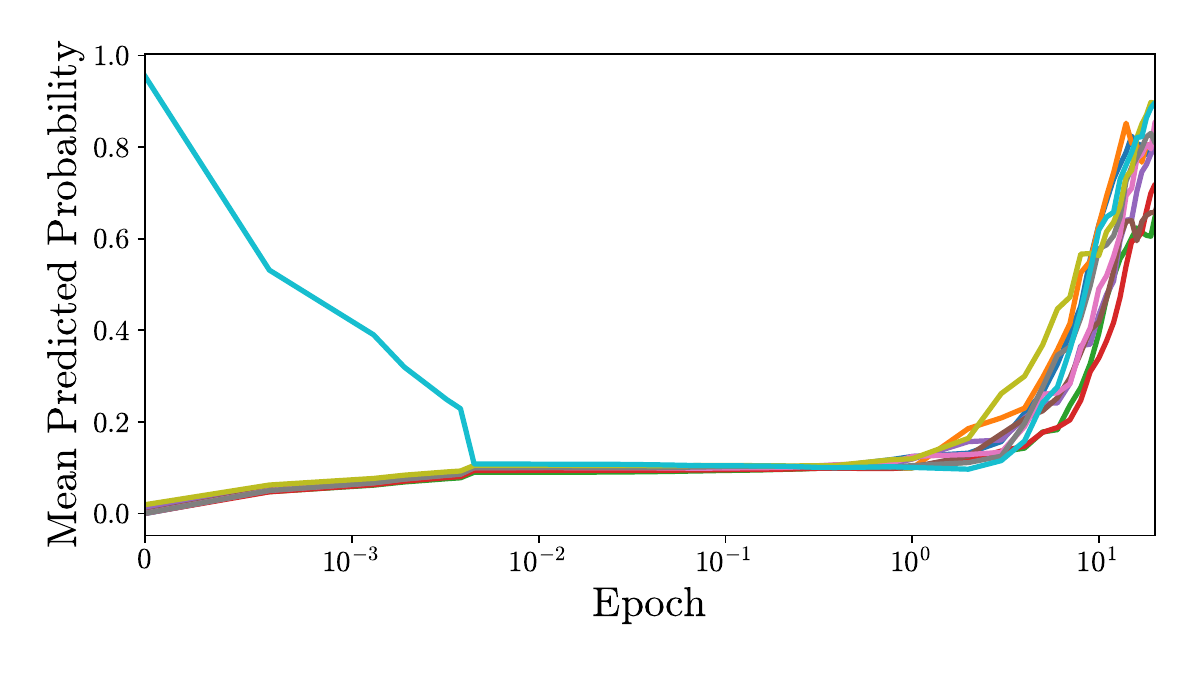}
    \end{subfigure}
    \hfill
    \begin{subfigure}[b]{0.49\textwidth}
        \centering
        \includegraphics[width=\linewidth]{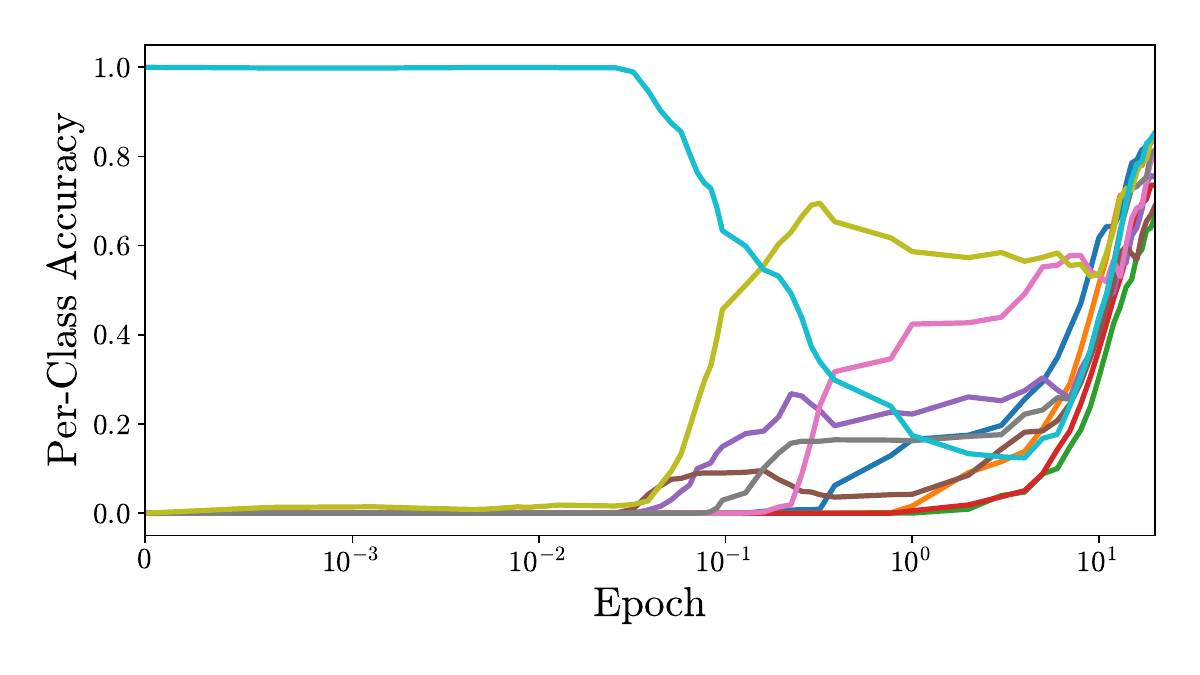}
    \end{subfigure}
\end{minipage}\\
% ---------- Row 2 ----------
BL &
\begin{minipage}{0.9\textwidth}
    \centering
    \begin{subfigure}[b]{0.49\textwidth}
        \centering
        \includegraphics[width=\linewidth]{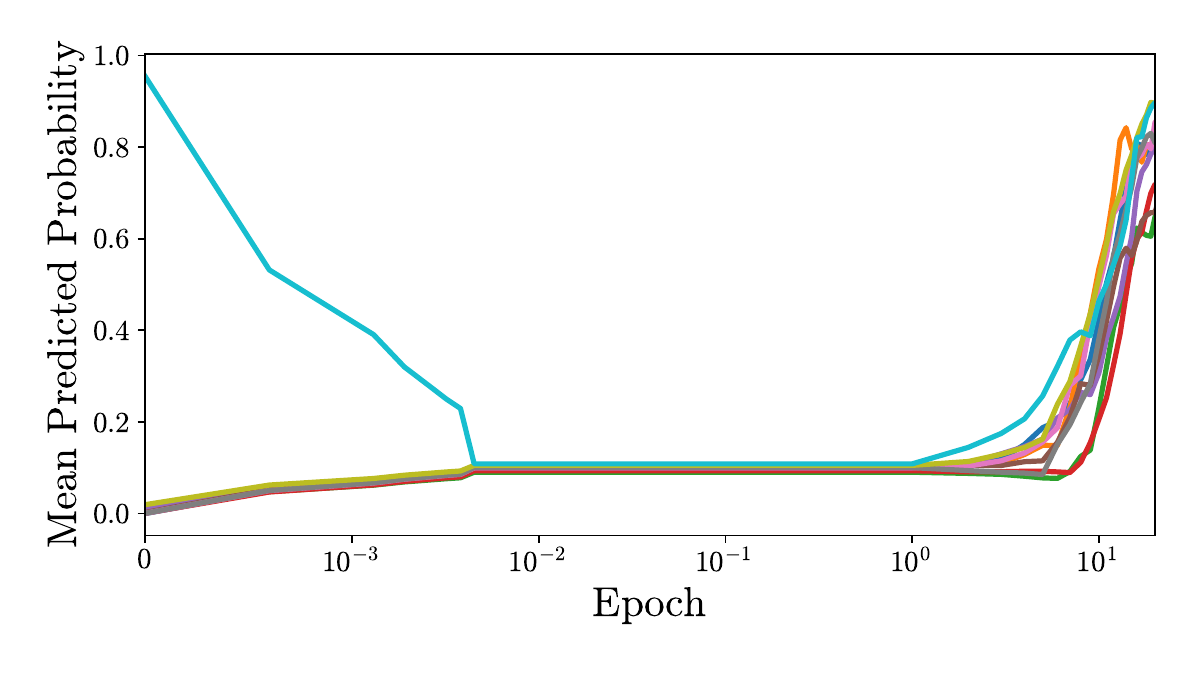}
    \end{subfigure}
    \hfill
    \begin{subfigure}[b]{0.49\textwidth}
        \centering
        \includegraphics[width=\linewidth]{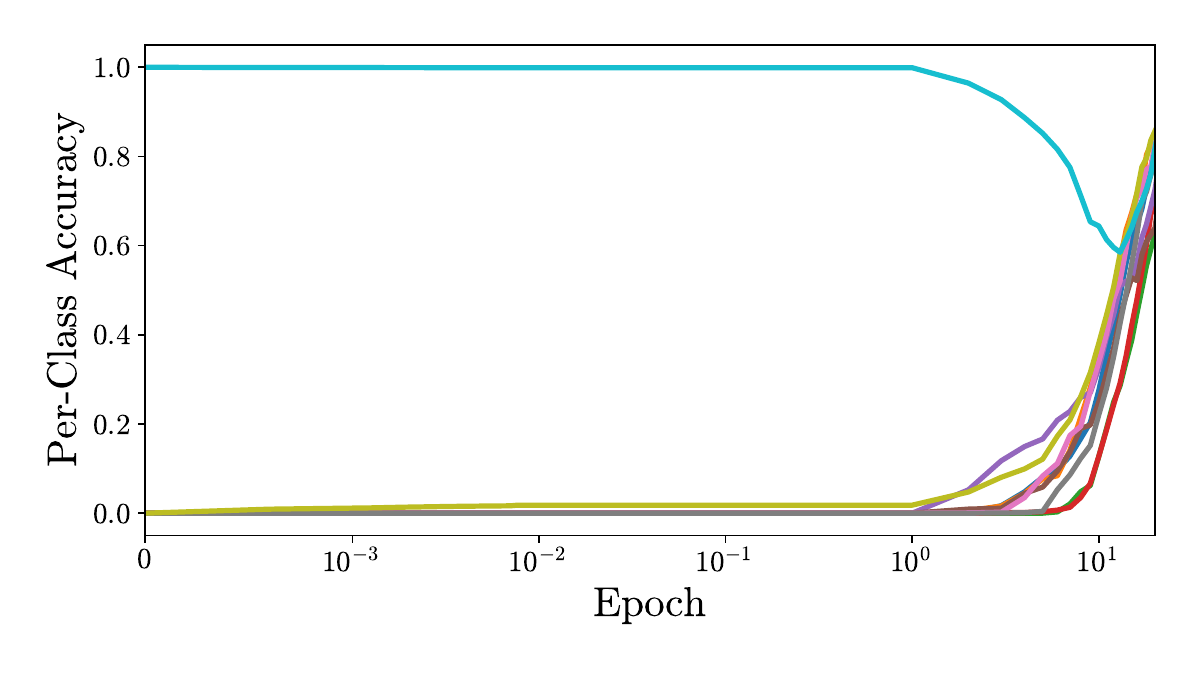}
    \end{subfigure}
\end{minipage}\\
% ---------- Row 3 ----------
PZ &
\begin{minipage}{0.9\textwidth}
    \centering
    \begin{subfigure}[b]{0.49\textwidth}
        \centering
        \includegraphics[width=\linewidth]{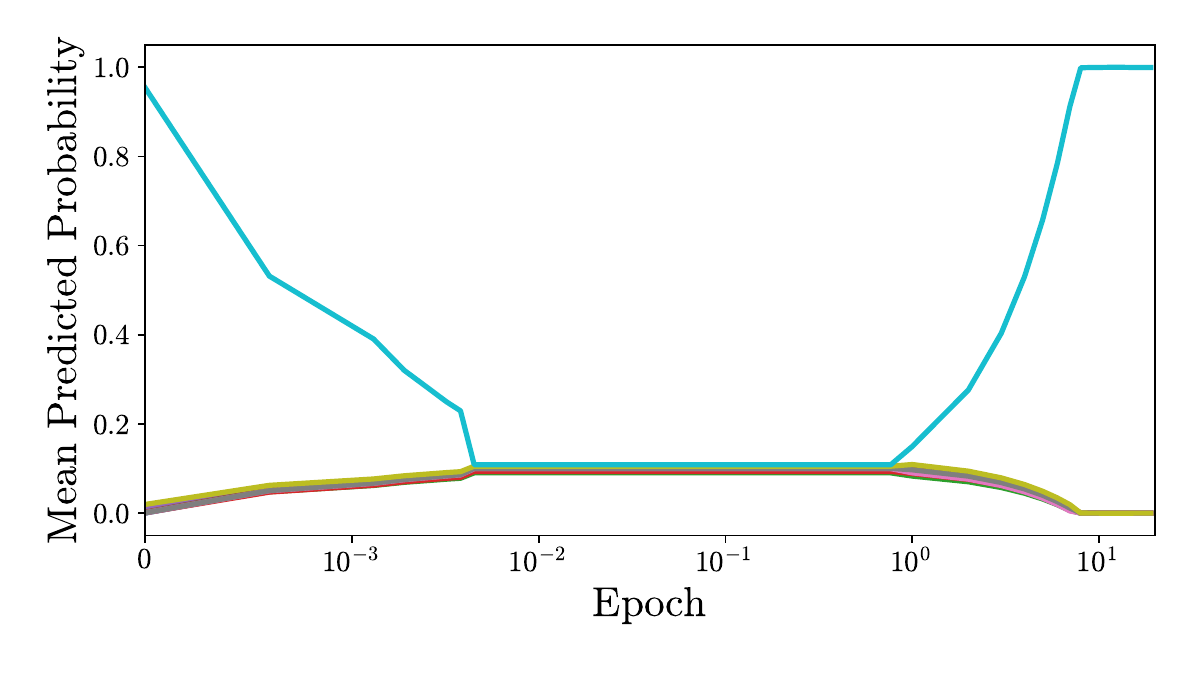}
    \end{subfigure}
    \hfill
    \begin{subfigure}[b]{0.49\textwidth}
        \centering
        \includegraphics[width=\linewidth]{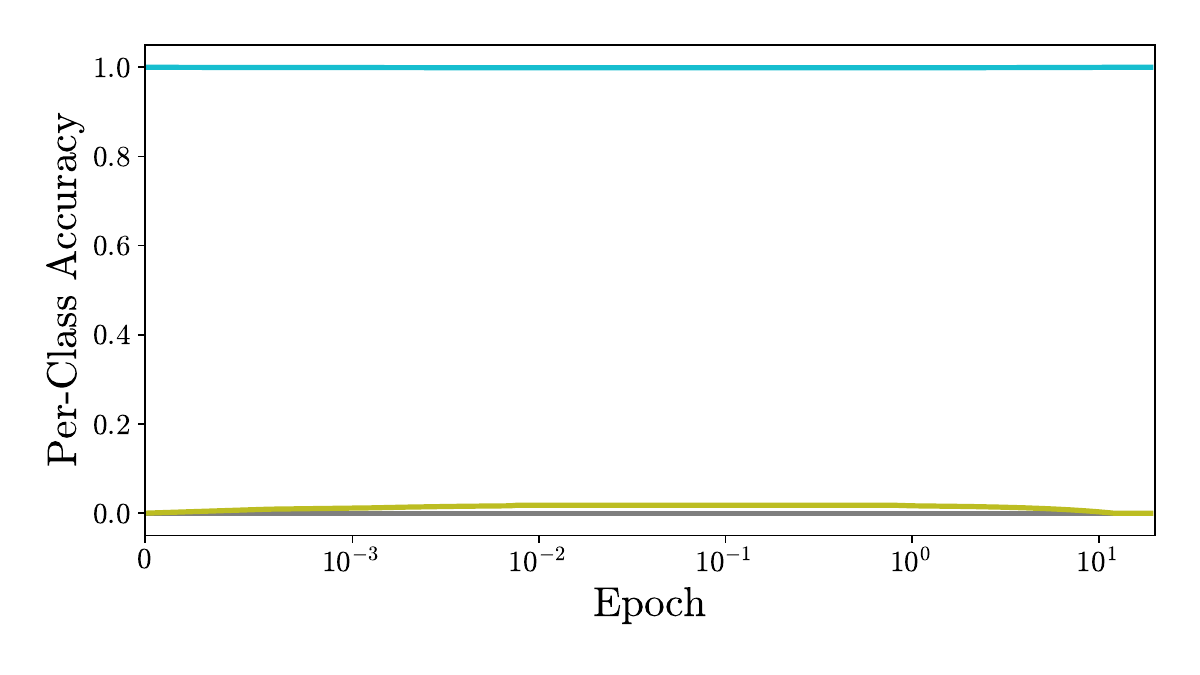}
    \end{subfigure}
\end{minipage}

\end{tabular}

\caption{Training dynamics for three choices of loss function: Cross-Entropy (CE; top row), Blurry Loss (BL; second row), and Piecewise-zero Loss (PZ; bottom row). Averaged Softmax probabilities (left column) and per-class accuracy (right column) are shown throughout training. 
Training duration is plotted on a log-scale, starting at 0 (before training) and showing fractional parts of the first epoch (batches). 
At the outset, the IGB effect causes probabilities and accuracies to be highly distinct between favoured and unfavoured classes for all loss functions (see \Cref{fig:mean_pred_probs}); however, as training progresses differences emerge. 
In all cases, predicted probabilities move towards each other during the first epoch and converge roughly to $p_y=0.1$ (approximating the class distribution of the dataset). 
After the first epoch, differences resulting from the choice of loss function reveal themselves. 
For Cross-Entropy, the predicted probabilities move as a group, gradually rising, with corresponding rises in per-class accuracies. 
Blurry loss behaves fairly similarly to Cross-Enropy, but with slightly slower dynamics and with accuracies lagging. 
For Piecewise-zero loss, predicted probability for the originally favoured class remains above all others, and again rises, while all others fall. 
In this case, the resulting per-class accuracies remain largely unchanged.}
\label{fig:training_dyn}
\end{figure*}

\section{Method}
\label{sec:method}

The interaction between Initial Guessing Bias and different loss functions are studied by examining the early training dynamics for a standard convolutional architecture on a simple, well-understood dataset. 
The experiments use a ResNet-50~\cite{he2016deep}, modified so that the first convolutional layer accepts single-channel input, with Kaiming He initialization~\cite{he2015delving}, the current standard in machine learning and implemented as default in PyTorch.
Training is performed on the CIFAR-10~\cite{krizhevsky2009learning} dataset. 
Three loss functions are considered: standard Cross-Entropy (CE)~\cite{good1952rational}, Blurry Loss (BL), and Piecewise-Zero Loss (PZ)~\cite{pellegrino2024loss}. 
All runs use the same method of random initialization and identical training settings (batch size of 32, using Stochastic Gradient Descent optimization~\cite{bottou2010large}, and 20 training epochs) to isolate the effect of the loss function.
PZ loss uses a cut-off parameter setting of $c=0.1$.
BL loss uses a weighting parameter setting of $\gamma=0.7$.

Prior to studying training dynamics, the presence and severity of IGB is characterised for the ResNet-50~\cite{he2016deep} architecture by visualizing the initial predicted probability distributions. 

To characterize early training behaviour, two quantities are recorded:
\begin{enumerate}[nosep]
    \item \textbf{Average Predicted Probability, $\bar{p}_k$}: \\
    Softmax probabilities, \mbox{$p(y=k|x)$}, are averaged over samples of each class, $k$, to track comprehensive model behaviour throughout training.
    % Averaged Softmax probabilities, $\bar{p}_k$, for each class, $k$, track overall model behaviour throughout training.
    % , ignoring whether data are labelled as each class.
    Samples of a given class, $k$, form a class subset $S_k = \{(x_i,y_i) | y_i = k\}$, such that the averaged Softmax probability is calculated as
    \begin{equation}
        \bar{p}_k=\frac{1}{|S_k|}\sum_{(x_i,y_i)\in S_k}{p(y_i=k|x_i)}.
    \end{equation}
    \item \textbf{Per-class Accuracy, $\mathrm{Acc}_k$}: \\
    Validation accuracy, computed separately for each class, $k$, tracks model performance relative to each class and is calculated as
    \begin{equation}
        \mathrm{Acc}_k=\frac{1}{|S_k|}\sum_{(x_i,y_i)\in S_k}{\mathds{1}\{\argmax_j p(j|x_i) = k\}}.
    \end{equation}
\end{enumerate}
These metrics are computed directly after model initialization and then during training at the end of every batch during the first epoch and at the end of each epoch thereafter.

\section{Results and Discussion}
\label{sec:results}

Directly after initializing the ResNet-50 model, predicted probabilities were collected and averaged over the validation set of CIFAR-10, demonstrating the severity of the Initial Guessing Bias in \Cref{fig:mean_pred_probs}. 
Observe that one class (class 0, randomly) is \textit{highly} favoured ($\bar{p}_0\approx1$) relative to other classes ($\bar{p}_y\approx0$ for $y\ne0$) indicating a high degree of bias.

\Cref{fig:training_dyn} shows the early-stage training dynamics for the three loss functions explored. 
Results for each loss function are plotted on separate rows, with predicted probabilities and per-class accuracy shown in the left and right columns, respectively.
Note that training duration is plotted on a log-scale, starting at 0 (before training) and showing fractional parts of the first epoch (batches). 

During the first epoch (before the $10^0$ tick mark), predicted probabilities tend towards $p_y=0.1$; however, per-class accuracy lags, remaining unchanged for longer, as correct classification occurs only for samples $(x_i,y_i)$ in which $p(y_i|x_i)$ and is maximal (uncommon for $\bar{p}_k\approx0.1$).
Because Blurry and Piecewise-zero loss de-emphasize samples with low $p_y$, in this regime, very few samples have any sizable influence on model training. 
In contrast, Cross-Entropy is most sensitive to samples with low $p_y$, leading to earlier increases in predicted probabilities accuracy across all classes.
While Blurry loss lags behind Cross-Entropy, for Piecewise-zero Loss there is insufficient training signal from samples of the non-favoured class, and training is almost entirely driven by the favoured class, leading to its rebound in predicted probability and constant perfect accuracy (while accuracy for all other classes remains poor).

Note that while the two robust loss functions de-emphasize (or entirely zero) the impacts of samples with low $p_y$, the training signal from samples (even just from one class) with high $p_y$ may steer the model towards increased $p_k$ for all classes. 

\section{Conclusion}
\label{sec:conclusion}

Initial Guessing Bias strongly influences early-stage training, and the choice of loss function plays a key role in how the model responds to this bias.
Cross-Entropy quickly reduces the imbalance and improves accuracy across all classes, while Blurry Loss achieves similar effects more slowly. 
Piecewise-Zero Loss provides too little signal for classes with low predicted probability, allowing the initially favoured class to dominate and preventing meaningful learning from the other classes. 
These findings underscore the importance of accounting for initialization biases and exercising care when selecting loss functions and other components of the training scheme.

\section*{Acknowledgments} % Turn off during review, on again later??? 
This research was enabled in part by support provided by Calcul Québec (\small\url{calculquebec.ca}) and the Digital Research Alliance of Canada (\small\url{alliancecan.ca}).

We acknowledge the support of the Government of Canada’s New Frontiers in Research Fund (NFRF), [NFRFT-2020-00073], and the support of the Natural Sciences and Engineering Research Council of Canada (NSERC) via \mbox{NSERC-CGS D}.

Nous remercions le Fonds Nouvelles Frontières en Recherche du gouvernement du Canada de son soutien (FNFR), [FNFRT-2020-00073], et le soutien du Conseil de Recherches en Sciences Naturelles et en Génie du Canada (CRSNG), CRSNG-BESC D.
\begin{figure}[h]
    \centering
    \includegraphics[width=\columnwidth]{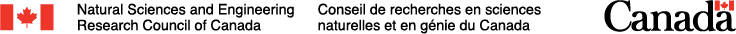}
\end{figure}

{
    \small
    \bibliographystyle{ieeetr}
    \bibliography{main}
}

\end{document}